\pdfoutput=1

\documentclass[11pt]{article}

\usepackage{acl}

\usepackage{times}
\usepackage{latexsym}

\usepackage[T1]{fontenc}

\usepackage[utf8]{inputenc}

\usepackage{microtype}
\usepackage{graphicx} 
\usepackage{comment}
\usepackage{multirow}
\usepackage{arydshln} 
\usepackage{subfigure}
\usepackage{amsmath} 
\usepackage{booktabs}
\usepackage{dirtytalk}
\usepackage{tablefootnote}

\title{A Generative User Simulator with GPT-based Architecture and Goal State Tracking for Reinforced Multi-Domain Dialog Systems}

\author{Hong Liu$^{1,3}$, Yucheng Cai$^{1,3}$, Zhijian Ou$^{1,3}$\thanks{~~Corresponding author. The code is released at \url{https://github.com/thu-spmi/GUS}}, Yi Huang$^{2,3}$, Junlan Feng$^{2,3}$ \\
  $^{1}$Speech Processing and Machine Intelligence (SPMI) Lab, Tsinghua University, Beijing, China \\
  $^{2}$China Mobile Research Institute, Beijing, China \\
  $^{3}$Tsinghua University-China Mobile Communications Group Co., Ltd. Joint Institute, Beijing, China \\
  \texttt{\{liuhong21,cyc22\}@mails.tsinghua.edu.cn},\\
  \texttt{ozj@tsinghua.edu.cn},\\
  \texttt{\{huangyi,fengjunlan\}@chinamobile.com}
}

\newcommand{\modelname}{GUS}
\begin{document}
\maketitle
\begin{abstract}
    Building user simulators (USs) for reinforcement learning (RL) of task-oriented dialog systems (DSs) has gained more and more attention, which, however, still faces several fundamental challenges.
First, it is unclear whether we can leverage pretrained language models to design, for example, GPT-2 based USs, to catch up and interact with the recently advanced GPT-2 based DSs.
Second, an important ingredient in a US is that the user goal can be effectively incorporated and tracked; but how to flexibly integrate goal state tracking and develop an end-to-end trainable US for multi-domains has remained to be a challenge.
In this work, we propose a generative user simulator (GUS) with GPT-2 based architecture and goal state tracking towards addressing the above two challenges.
Extensive experiments are conducted on MultiWOZ2.1.
Different DSs are trained via RL with GUS, the classic agenda-based user simulator (ABUS) and other ablation simulators respectively, and are compared for cross-model evaluation, corpus-based evaluation and human evaluation.
The GUS achieves superior results in all three evaluation tasks.

\end{abstract}

\section{Introduction}
\label{sec:introduction}
Task-oriented dialog (TOD) systems are mainly designed to help users accomplish specific goals, such as finding restaurants or booking flights.
The dialog system (DS) usually consists of several modules - dialog state tracking (DST), database querying (DB), dialog policy (DP) and natural language generation (NLG).
Recent studies recast these modules all as conditional generation of tokens and build on some pretrained language model (PLM) such as GPT-2 \cite{radford2019gpt2} as the backbone.
Fine-tuning PLM over annotated dialog datasets via supervised learning (SL) has shown state-of-the-art results \cite{hosseini2020simple, peng2020etal, kulhanek2021augpt, yang2021ubar, lee-2021-improving-end}, thanks to the powerful generation ability of PLMs.

\begin{figure}[t]
\centering
	\includegraphics[width=0.95\linewidth]{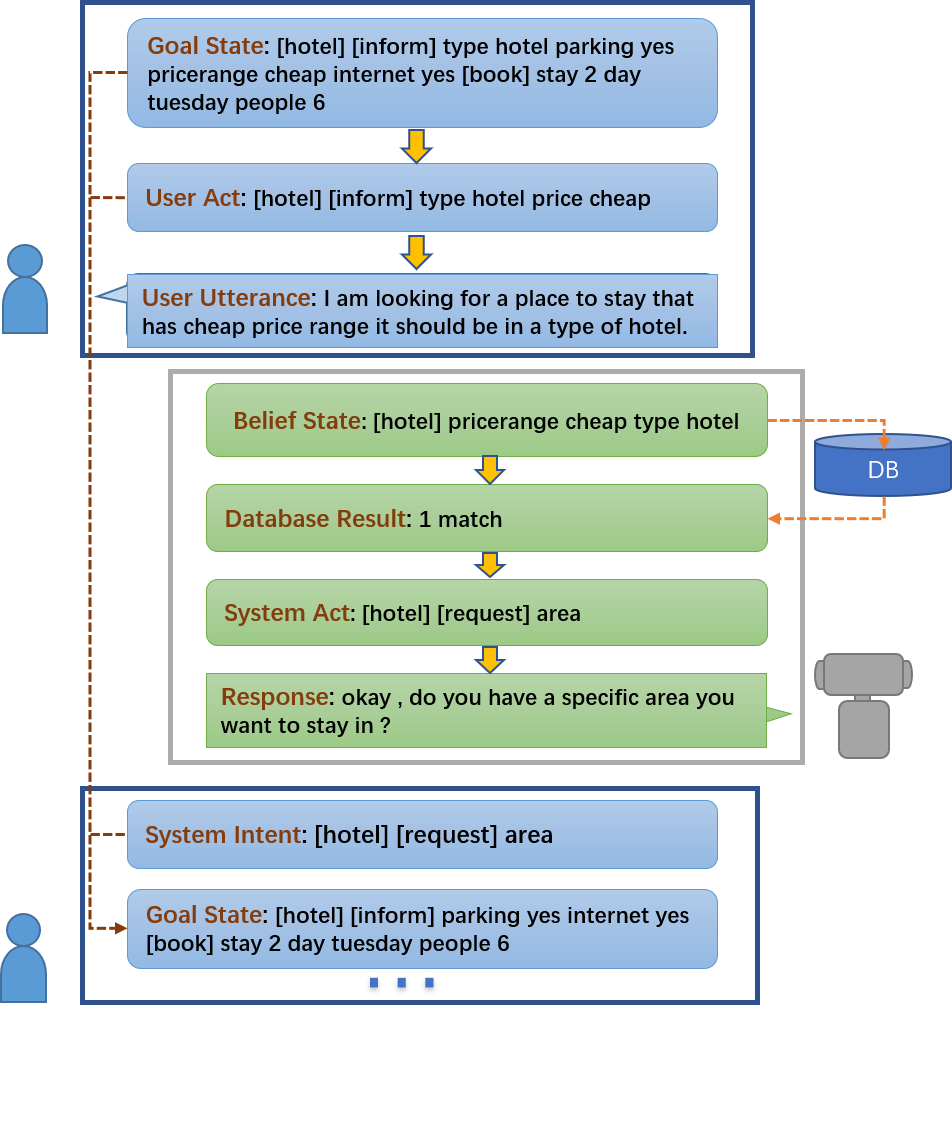}
	\vspace{-2em}
	\caption{The information flow in a task-oriented dialog. Domains and intents are enclosed by square brackets.}
	\vspace{-1em}
	\label{fig:flow}
\end{figure}

However, supervised trained agents could become biased by the annotations, and it has long been recognized that reinforcement learning (RL) could be applied to policy learning for the agent \cite{young2013pomdp}, which aims at goal-directed learning from interaction between the dialog agent and the user. Interaction with human users is expensive and time-consuming in practice. Therefore, an alternative approach, building user simulators (USs), has gained more and more attention, which, however, still faces several fundamental challenges.

First, note that the recent research on building dialog agents has been significantly advanced with the end-to-end trainable generative approach based on PLMs such as GPT-2.
However, prior work on user simulators are mostly LSTM-based, not utilizing any PLMs, as reviewed in Table \ref{tab:comparison}. It is unclear whether we can leverage PLMs to design, for example, GPT-2 based\footnote{It can be seen that the discussion and the proposed method in the remainder of this paper can also be applied to other PLMs such as T5 \cite{raffel2020t5}, not limited to GPT-2.} user simulators, to catch up and interact with the GPT-2 based dialog agents. This has not ever been systematically examined, to the best of our knowledge.
We leave detailed discussion to Related Work section, where we review prior work on USs from a number of important features in building USs.


Second, an important ingredient in a US is that the user goal can be incorporated and tracked.
Task-oriented dialog systems are characterized by a user goal, which is composed of user constraints and requests.
The user goal ensures that the user behaves in a consistent, goal-directed manner, and the system agent is considered successful if it is able to fulfill the user goal by the end of a dialog session.
Thus, it is desirable for the US to track the completion process of the goal explicitly (which we call goal state tracking in this paper), as did in the classic agenda-based user simulator (ABUS) \cite{schatzmann-etal-2007-agenda}.
However, the goal state tracking process is overlooked in later data-driven USs \cite{asri2016sequence, GurHUS, papangelis-etal-2019-collaborative}, or realized by binary vectors \cite{kreyssig-etal-2018-neural,lin-etal-2021-domain,tseng-etal-2021-transferable}, or only works at the semantic level \cite{takanobu-etal-2020-multi}.
How to flexibly integrate goal state tracking and develop an end-to-end trainable US for multi-domains has remained to be a challenge.

In this work, we propose a generative user simulator (GUS) with GPT-2 based architecture and goal state tracking towards addressing the above two challenges in building end-to-end trainable USs for reinforced multi-domain dialog systems.
Basically, a US, interacting with a DS in natural languages, needs several modules - natural language understanding (NLU) of system responses,  goal state tracking (GST) to refresh the remained constrains and requests that need to send subsequently, user policy (UP), and natural language generation (NLG).
The information flow in a task-oriented dialog between a US and a DS is illustrated in Figure \ref{fig:flow}.
In generative user simulator (GUS), we recast these modules in US all as conditional generation of tokens, similar to the recent approach of finetuning PLMs such as GPT-2 to build end-to-end trainable generative DSs.

\begin{figure}[t!]
	\centering
	\subfigure[Architecture of Dialog System (DS)]
	{\label{fig:attention_a}
	\includegraphics[width=\columnwidth]{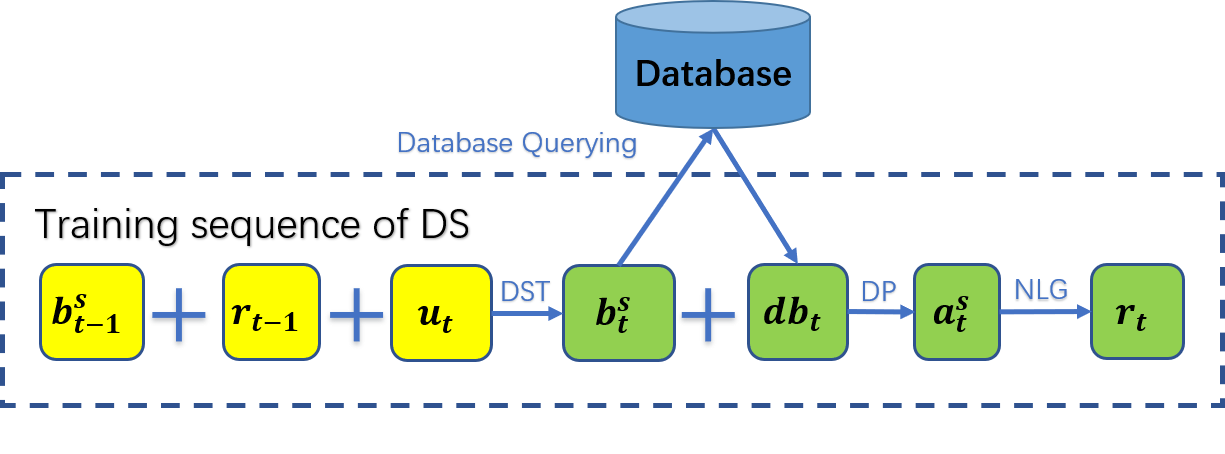}}
	\subfigure[Architecture of User Simulator (US)]
	{\label{fig:attention_r}
	\includegraphics[width=0.8\columnwidth]{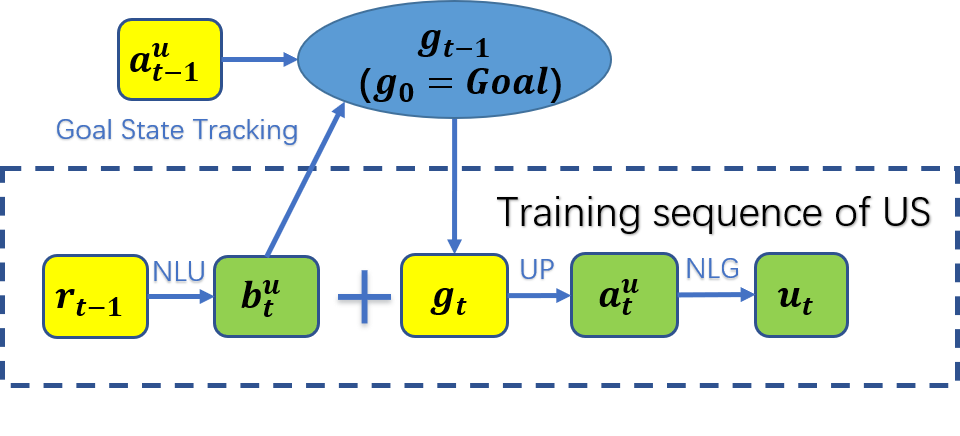} }
	\caption{The generative architecture of dialog system and user simulator in our experiments.Yellow boxes represent the conditioning input of the model during generation, and green boxes the targeting output.}
	\vspace{-0.8em}
	\label{fig:structure}
\end{figure}

To be specific in this paper, we use the GPT-2 based architecture for GUS to generate user acts and user utterances, and constantly track the goal state according to the user acts and system acts of the previous turn, which is shown in Figure \ref{fig:structure}.
In this work, the definition of goal state is similar to the agenda in ABUS \cite{schatzmann-etal-2007-agenda}, which represents a collection of pending user acts that are needed to elicit the information specified in the goal. The maintenance of goal state includes not only removing the completed user acts, but also changing the user goal when the system cannot find a requested entity. 


Extensive experiments are conducted on MultiWOZ2.1 \cite{eric2019multiwoz}. Different DSs are trained via RL with GUS, ABUS and other ablation simulators respectively, and are compared for cross-model evaluation, corpus-based evaluation and human evaluation.
The GUS achieves superior results in all three evaluation tasks.

\begin{table}[t]
\centering
\resizebox{\linewidth}{!}{
\begin{tabular}{l|c|c|c|c|c|c}
\multirow{2}{*}{US} &\multirow{2}{*}{PLM} &Goal State & Cross-model &Compared &Natural Lang. &Multi- \\
& &Tracking &Evaluation &with DS-SL &Interaction &Domain\\
\hline
\citet{schatzmann-etal-2007-agenda} &N &Y &N &N &N &N \\
\citet{asri2016sequence} &N &N &N &N &N &N \\
\citet{DBLP:conf/asru/LiuL17} &N &N &N &Y &Y &N\\
\citet{GurHUS} &N &N &N &N &N &N \\
\citet{kreyssig-etal-2018-neural} &N &Y &Y &N &Y &N\\
\citet{papangelis-etal-2019-collaborative} &N &N &N &Y &Y &N\\
\citet{shi-etal-2019-build} &N &N &Y &N &Y &N\\
\citet{takanobu-etal-2020-multi} &N &Y &N &Y &N &Y\\
\citet{lin-etal-2021-domain} &N &Y &Y &N &N &Y\\
\citet{tseng-etal-2021-transferable} &N &Y &N &Y &Y &Y\\
\hline
GUS &Y &Y &Y &Y &Y &Y\\
\end{tabular}
}
\caption{Comparison of prior different user simulators from a number of important features in building USs. DS-SL denotes dialog system (DS) trained by supervised learning (SL). See Section \ref{sec:related} for detailed meaning of each feature by column.}
\vspace{-0.5em}
\label{tab:comparison}
\end{table}

\section{Related Work}
\label{sec:related}

\paragraph{Novelty}
In Table \ref{tab:comparison}, we review prior work on USs from a number of important features in building USs, including whether or not the US is based on any PLMs, the US conducts goal state tracking, the cross-model evaluation \cite{schatztnann2005effects} is conducted to assess the performance of the US, the DS trained via RL with the US is compared to the DS trained via supervised learning, the US and the DS interact in natural languages \footnote{This means that during reinforcement training of the DS with the US, the US accepts the system response in natural language. In contrast, for those USs with `N' marked in the `Natural Lang. Interaction' column, the system acts are directly fed to the US so that the US does not need a natural language understanding module. For such as case, the US is also said to work at the semantic level.}, the US is designed to work for multi-domain dialogs. It is clear from Table \ref{tab:comparison} that our proposed GUS is distinctive, which represents the first US that possesses all these desirable features, to the best of our knowledge. More discussions are provided in the following.

\paragraph{US Architecture}
A variety of user simulators have been studied, either rule-based or data-driven. 
A classic rule-based US is the agenda-based user simulator (ABUS) \cite{schatzmann-etal-2007-agenda}.
Different data-driven US models are proposed with different architectures and characteristics.
\citet{asri2016sequence} develops a LSTM-based seq2seq US on the single-domain DSTC2 dataset and generates semantic-level user acts. 
\citet{GurHUS} proposes a GRU-based hierarchical seq2seq framework for US (HUS) and further introduces a latent variable to control the diversity of dialogue (VHUS).
NUS \cite{kreyssig-etal-2018-neural} extracts feature vectors related to current goal states and feeds to a LSTM seq2seq model to output natural languages.
\citet{shi-etal-2019-build} make extensive comparisons for six user simulators, based on two user policy modules (seq2seq or agenda based) and three NLG modules (template, retrieval or seq2seq).
TUS in \cite{lin-etal-2021-domain} designs domain-independent features and implements the user policy as multi-class classification so that TUS could be easily adapted to new domains.
Some studies aim to jointly optimize DS and US. The USs used in these studies are mostly based on LSTM seq2seq architectures \cite{DBLP:conf/asru/LiuL17,papangelis-etal-2019-collaborative,tseng-etal-2021-transferable}, or simply as multi-class classification for action selection with feed-forward networks \cite{takanobu-etal-2020-multi}.

\paragraph{Goal State Tracking in US}
ABUS is classic in goal state tracking, where the pending user acts are tracked in a stack-like structure, called agenda.
ABUS is rule-based, generating user acts by pushing and popping hand-crafted rules from agenda.
The goal state tracking process is overlooked in some later studies of data-driven USs \cite{asri2016sequence, GurHUS, papangelis-etal-2019-collaborative}, where the US is always conditioned on the whole initial user goal at each turn.
Some data-driven USs explicitly track goal states but employ binary vectors \cite{kreyssig-etal-2018-neural,lin-etal-2021-domain,tseng-etal-2021-transferable}.
The US in \cite{takanobu-etal-2020-multi} represents goal states by tokens, which is flexible, but the US only interacts with the DS at the semantic level (not end-to-end trainable).

\section{Preliminaries}
\label{sec:background}
\paragraph{Notations} According to the information flow in a task-oriented dialog between a US and a DS as illustrated in Figure \ref{fig:flow}, we let $g_t$ denote the user goal state, $a_t^u$ the user act, $u_t$ the user utterance, $b_t^s$ the system belief state, $db_t$ the database result, $a_t^s$ the system act, and $r_t$ the system response, respectively, at turn $t=1,\cdots,T$, for a dialog of $T$ turns. 
Moreover, in this paper we are interested in building end-to-end trainable US that can interact with the DS in natural languages. Thus, we introduce a NLU module in the US, which takes the system response $r_{t}$ as input and infer system intent. The NLU result is denoted by $b_t^u$, or loosely speaking, referred to as the user belief state. Notably, the US belief state $b_t^u$ denotes the US's understanding only about the previous system response, and accordingly is labeled as $a_{t-1}^s$ in training.
$b_t^u$ is not of accumulated nature, since the US uses the goal state $g_t$ to summarize the dialog history encountered by the US\footnote{In contrast, the system belief state $b_t^s$ summarizes the dialog history encountered by the DS. This subtle difference makes sense, since the roles of the DS and US are different.}.

\paragraph{GPT-2-based Generative Architecture} 
In this work, all variables defined in the last paragraph for the US and DS are converted to token sequences, like in DAMD \cite{zhang-etal-2020-probabilistic}.
So pretrained language models (LMs) such as GPT-2 can be finetuned to build end-to-end trainable DS and US, as will be introduced later.
To be clear, GPT-2 \cite{radford2019gpt2} in this paper refers to the particular class of causal LMs, which computes conditional probabilities for next-token generation via self-attention based Transformer neural network \cite{vaswani2017attention}.
Given a particular form of conditional model, $p(output|input)$, where $input$ and $output$ are token sequences, the GPT-2 model can be finetuned over training samples $(input, output)$ (often referred to as training sequences \cite{hosseini2020simple}), and after finetuning, the model can be used for generation, i.e., generating $output$ after receiving $input$.

\paragraph{Generative Dialog System} The main task for a dialog system (DS) is, for each dialog turn $t$, to generate (or say, predict)\footnote{Note that database result $db_t$ is deterministically obtained by querying database using the predicted $b_t^s$. We omit $db_t$ in the discussion for simplicity.} $b_t^s$, $a_t^s$ and $r_t$, given $u_t$ and dialog history $u_1, r_1, \cdots, u_{t-1}, r_{t-1}$.
A recent progress in building DS is that all variables are represented by token sequences, and the workflow of a dialog system (DST, DP and NLG) is unified into a single sequence generation problem, which can be accomplished by a causal LM such as GPT-2 \cite{hosseini2020simple,Liu2022RevisitingMG}. 
In this paper, we employ the Markov generative architecture (MGA) for DS, which is introduced in \citet{Liu2022RevisitingMG} and shows efficiency advantages in memory, computation and learning over non-Markov DS models like SimpleTOD \cite{hosseini2020simple}.
Specifically, for DS to predict $b_t^s$, $a_t^s$ and $r_t$ at each turn $t$, we use only the belief state $b_{t-1}$ and response $r_{t-1}$ from previous turn along with current user utterance $u_t$, as shown in Figure \ref{fig:structure}(a). The DS can thus be trained via finetuning GPT-2 by maximizing the following conditional likelihood over labeled training sequences for supervised learning (SL):
\begin{equation}
\label{eq:sup-ds}
\begin{aligned}
    \mathcal{J}_{\text{DS-SL}} &=\log p_{\theta}(b_t^s, a_t^s, r_t|b_{t-1}^s, r_{t-1}, u_t)\\
    &= \sum_{i=1}^{|b_t^s \oplus a_t^s \oplus r_t|} \log p_\theta(c_i| b_{t-1}^s, r_{t-1}, u_t, c_{<i}) 
\end{aligned}
\end{equation}
where $\oplus$ denotes the concatenation of sequences,
$|b_t^s \oplus a_t^s \oplus r_t|$ denotes the length in tokens, and $c_i$ denotes the $i$-th token. The DS parameters are actually a set of GPT-2 parameters, collectively denoted by $\theta$.

\section{Method: Generative User Simulator}
\label{sec:method}

An end-to-end trainable US needs several modules - natural language understanding (NLU) of system responses,  goal state tracking (GST), user policy (UP), and natural language generation (NLG).
Inspired by the recent approach of finetuning PLMs such as GPT-2 to build end-to-end trainable generative DSs, we propose an end-to-end trainable generative user simulator (GUS), which generally refer to the approach of recasting all the modules in the US (NLU, UP, and NLG) as conditional generation of tokens based on finetuning PLMs such as GPT-2.
In the following, we first introduce the GUS model including goal state tracking and GPT-2 based architecture. Then, we describe how GUS is trained and used for reinforcement training of the DS.

\subsection{GUS Model}
\paragraph{Goal State Definition}
Crucially, the interaction between the user and the system is motivated by the user goal, which is composed of user constraints and requests such as booking a cheap hotel. The goal state, in this paper, is defined as the uncompleted part of the user goal at each turn.
Similar to \citet{kreyssig-etal-2018-neural}, we accumulate the annotated user acts backwards turn by turn to obtain the goal state annotation at each turn. The accumulation process is illustrated in Appendix~\ref{sec:processing}.
The goal state obtained at the first turn corresponds to the initial user goal for the whole dialog session. 

\paragraph{Goal State Tracking} 
Given the goal state annotations at each turn, the US can be trained via teacher-forcing to mimic the user behaviors.
When the US is applied to interact with the DS for evaluation or for reinforcement training of the DS, the US needs to track the completion process of the goal to update the goal state turn by turn, which we call goal state tracking.
There are three types of user intents in the goal state $g_t$ - \emph{inform, book and request}. 
The slots and values for the first two types of intents in $g_t$ are denoted by $g_t^{\emph{constraint}}$ and those of the \emph{request} intent as $g_t^{\emph{request}}$. The update rule of $g_t$ at turn $t$ is designed to be as follows:
\begin{equation}
\label{eq:update}
\begin{aligned}
    &g_t^{\emph{constraint}}=g_{t-1}^{\emph{constraint}} \ominus a_{t-1}^{u,\emph{inform}}\\
    &g_t^{\emph{request}}=g_{t-1}^{\emph{request}} \ominus b_t^{u,\emph{inform}}
\end{aligned}
\end{equation}
where $a_{t-1}^{u,\emph{inform}}$, $b_t^{u,\emph{inform}}$ are the informable slots and values in user act $a_{t-1}^{u}$ and user belief state $b_{t}^{u}$ respectively and $\ominus$ denotes removing the corresponding slots and values.
Moreover, the slot values in the initial user goal may be changed during the interaction (i.e., goal change). When the DS expresses \emph{no-offer} intent, which means no entities in the database satisfy the constraints of the goal, we randomly select one slot in the \emph{no-offer} intent and replace its value with another value in the ontology. 

\paragraph{GPT-2-based Architecture} 
\label{sec:US}
The main task for a US is, conditional on the user goal, to iteratively understand the system response, track goal state, decide user act, and generate user utterance. 
In this work, we find that the recent approach of finetuning GPT-2 for conditional generation can be similarly applied to build US.
Specifically, we employ Markov generative architecture \citep{Liu2022RevisitingMG}. The US is designed to firstly infer the system intent, i.e., user belief state $b_t^u$ of turn $t$ from the previous system response $r_{t-1}$, which could be modeled as $p_\phi(b_t^u|r_{t-1})$. After obtaining $b_t^u$, the goal state will be updated according to the rule in Eq.~\eqref{eq:update}. Then, the US will generate user act and user utterance sequentially conditioned on the previous system response, user belief state, and the updated goal state. The resulting US is called GUS and could be modeled as $p_\phi(a_t^u, u_t|r_{t-1}, b_t^u, g_t)$.
The GUS parameters are actually another set of GPT-2 parameters, collectively denoted by $\phi$.

\subsection{GUS Training}
The GUS model can thus be trained via finetuning GPT-2 by maximizing the following conditional likelihood over labeled training sequences for supervised learning (SL):
\begin{equation}
\label{eq:sup-us}
\begin{aligned}
    \mathcal{J}_{\text{US-SL}} &=\log p_\phi(b_t^u|r_{t-1})\\
    &+\log p_\phi(a_t^u, u_t|r_{t-1}, b_t^u, g_t)
\end{aligned}
\end{equation}
Note that during supervised learning, the user belief state $b_t^u$ is labeled by directly copying the system act $a_{t-1}^s$ from the previous turn.

\subsection{Reinforcement Optimization of DS through Interaction with US}
\label{sec:optimization}
\paragraph{RL Setup} 
The DS and US described above will first be trained using supervised learning with the objectives in Eq.~\eqref{eq:sup-ds} and Eq.~\eqref{eq:sup-us} respectively. After supervised learning, we can perform RL optimization on the DS through interactions with the US. 
The DS agent view the US as the environment and use its conditional model $p_{\theta}(b_t^s, a_t^s, r_t|b_{t-1}^s, r_{t-1}, u_t)$ as its policy. 
Here the policy of the DS involves generating not only system act $a_t^s$, but also belief state $b_t^s$ and system response $r_t$.
This is different from some previous studies of learning reinforced DS, e.g., \cite{DBLP:conf/asru/LiuL17,papangelis-etal-2019-collaborative,tseng-etal-2021-transferable}, which only use RL to optimize the selection of system acts (but all use traditional LSTM seq2seq architectures).
However, thanks to the representation power of GPT-2, recursively predict (or say, decide about) $b_t^s$, $a_t^s$ and $r_t$ in one policy yields the best performance in our experiment.
In Section~\ref{sec:policy}, we compare different schemes for policy definition for the DS agent with more discussions.

\paragraph{RL Optimization} 
We apply the policy gradient method \cite{sutton2000policy} to optimize the DS for RL. We first let the two agents interact with each other based on the user goals from the goal generator provided by ConvLab-2 \cite{zhu2020convlab2}. Then we calculate the reward $R_t$ for each turn, as detailed below. The return $U_{i,t}$ for the action of turn $t$ at the $i$-th step is $\gamma^{|A_t|-i}R_t$, where $\gamma$ is the discounting factor and $|A_t|$ is the policy sequence length of turn $t$. We update the DS with the following policy gradients:
\begin{equation}
\label{eq:rl-ds}
\begin{aligned}
    &\nabla_\theta \mathcal{J}_{\text{DS-RL}}=\sum_{i=1}^{|b_t^s \oplus a_t^s \oplus r_t|} U_{i,t} \nabla_\theta \log p_\theta(c_i)
\end{aligned}
\end{equation}
where $p_\theta(c_i)$ denotes $p_\theta(c_i| b_{t-1}^s, r_{t-1}, u_t, c_{<i})$.

\paragraph{Reward Settings} 
A number of different settings for reward have been studied, as described in the following. The three settings are separately tested, and the experimental results are given in Section~\ref{sec:reward}.\\
1) Success. If a dialog is successful, we set the reward of each turn to 1, otherwise it is set to be 0;\\
2) A turn-level synthetic reward similar to \citet{tseng-etal-2021-transferable, takanobu-etal-2020-multi}, which consists of requesting reward (+0.1 for each), repeating punishment (-0.5 for each) and task completion reward (the proportion of tasks completed) of the DS;\\
3) A Sigmoid synthetic reward obtained by mapping the synthetic reward to [0,1] interval using the Sigmoid function. This setting is designed to exclude the influence of the value range of reward because the value range is different between the Success reward and the synthetic reward. 

\section{Experiments}
\label{sec:exp}
\subsection{Dataset}
Experiments are conducted on MultiWOZ2.1 \cite{eric2019multiwoz}, which is an English multi-domain task-oriented dialog dataset of human-human conversations. It contains 10.4k dialogs, collected in a Wizard-of-Oz setup over seven domains. The dataset contains the annotations of system belief state, system act, and user act for every turn.
\subsection{Evaluation Metrics}
\label{sec:metrics}
Evaluating the quality of a US is not trivial.
The performance of the reinforced DS trained with a specific US gives an \emph{indirect} assessment of the quality of the US.
Considering that a main purpose of developing USs is to help train RL based DSs, this indirect assessment makes sense and is widely employed \cite{kreyssig-etal-2018-neural,shi-etal-2019-build,lin-etal-2021-domain}.
We conduct both automatic evaluation and human evaluation of the DSs trained with different USs.
Additionally, we also ask human graders to \emph{directly} assess the performance of different USs, by reading and scoring the generated utterances from the USs.

\vspace{-0.5em}
\paragraph{Automatic Evaluation} It could be interaction-based or corpus-based. 
For both manners, we can calculate \emph{Inform} and \emph{Success} for measuring the performance of the DSs. 
\emph{Inform Rate} measures how often the entities provided by the system are correct. \emph{Success Rate} refers to how often the system is able to answer all the requested attributes by user. 
\emph{BLEU Score} is used to measure the fluency of the generated system responses when conducting corpus-based evaluation.

\paragraph{Human Evaluation} We conduct human evaluation, where human graders are recruited to assess the quality of dialogs generated by the US and the DS trained with it. Similar to \citet{su2021multitask}, for each dialog, the grader will score the conversation on a 3-point scale (0, 1, or 2)\footnote{Three scales (0, 1 and 2) denote three degrees - not at all, partially and completely, respectively.} by the following 3 metrics for the DS and 2 metrics for the US:
\begin{itemize}
    \setlength\itemsep{0.1em}
    \item Success. This metric measures if the DS successfully completes the user goal by interacting with the US;
    \item DS Coherency (DS-coh). This metric measures whether the system’s response is logically coherent with the dialogue context;
    \item DS Fluency (DS-Flu). This metric measures the fluency of the system’s response.
    \item US Coherency (US-Coh). This metric measures whether the simulator’s utterance is logically coherent with the dialogue context;
    \item US Fluency (US-Flu). This metric measures the fluency of the simulator’s utterance.    
\end{itemize}
\subsection{Baseline}
The DS model is as described in Section \ref{sec:background}.
We compare \modelname{} with the classic rule-based simulator ABUS \cite{schatzmann-etal-2007-agenda}. 
We use the simulator in the ConvLab-2 \cite{zhu2020convlab2} toolkit, which provides an instantiation of ABUS on MultiWOZ \cite{budzianowski2018large}, including BERT-based NLU and template-based NLG. The ABUS in ConvLab-2 has a goal generator module, which we use for driving the interaction between the DSs and the proposed \modelname{}.
Remarkably, the TUS paper \citep{lin-etal-2021-domain} has revealed the shortcoming of VHUS \citep{GurHUS}, which performs much worse than ABUS. Also, it is concluded that TUS has a comparable performance to the rule-based ABUS in cross-model evaluation. Thus, in this paper, we mainly compare GUS with ABUS, which is a very strong baseline.

\section{Main Results}
\subsection{Cross-Model Evaluation}
\label{sec:rl_results}
Cross-model evaluation is a type of automatic evaluation \cite{schatztnann2005effects} to compare different USs.
The main idea is that if the DS trained with a specific US performs well on all USs (not just on the one that the DS was trained with), it indicates the specific US with which the DS was trained is of good quality (realistic), and thus the DS is likely to perform better on real users.

Specifically, we first train a DS and a US separately on training data based on the supervised learning objectives described in Eq.~\eqref{eq:sup-ds} and Eq.~\eqref{eq:sup-us}. The resulting models are referred to as DS-SL and GUS respectively. Then we further optimize DS-SL by policy gradient in Eq.~\eqref{eq:rl-ds} on interaction with either ABUS or GUS, and obtain DS-ABUS and DS-GUS respectively.
For either of ABUS and GUS, RL trainings (all starting from DS-SL) are independently taken for three times with different random seeds.
Each specific DS model is then tested on both ABUS and GUS. We use the same 1000 randomly generated goals for each test. Further implementation details can be found in Appendix~\ref{sec:implementation}.
Table~\ref{tab:cross} shows the cross-model evaluation results\footnote{Similar tables to Table 2 have been used in previous work such as NUS \citep{kreyssig-etal-2018-neural} and TUS \citep{lin-etal-2021-domain}. A common practice of reading such tables is row-by-row comparison. This is exactly what the cross-model evaluation means.}.
\begin{table}[t]
\renewcommand\arraystretch{1.1}
\centering
\resizebox{\linewidth}{!}{
\begin{tabular}{l|cc|cc}
\hline
\multirow{2}{*}{DS $\backslash$ US} &\multicolumn{2}{c|}{ABUS} &\multicolumn{2}{c}{GUS}\\
&Inform &Success &Inform &Success\\
\hline
DS-SL &0.864 &0.791 &0.781 &0.736\\
\hline
DS-ABUS$_{best}$ &0.885 &0.816 &0.783 &0.741\\
DS-GUS$_{best}$ &0.881 &0.810 &0.864 &0.808\\
\hline
DS-ABUS$_{avg}$ &0.889 &0.793 &0.793 &0.735\\
DS-GUS$_{avg}$ &0.872 &0.801 &0.859 &0.802\\
\hline
\end{tabular}
}
\caption{Cross-model evaluation results. The subscripts $best$ and $avg$ denote the best and the average from 3 independent RL experiments with different random seeds.}
\label{tab:cross}
\end{table}

It can be seen from Table~\ref{tab:cross} that the DS trained with GUS (DS-GUS) performs well on both ABUS and GUS, while the DS trained with ABUS (DS-ABUS) only performs well on ABUS and achieves much lower Inform and Success when tested with GUS. This indicates the superiority of GUS over ABUS, being more helpful in training reinforced DSs that perform well on both USs.
Moreover, DS-GUS also outperforms the supervised baseline (DS-SL) on both USs. This shows the practical benefit brought by training DSs via RL on interaction with the proposed GUS.
Such comparison of RL and SL is overlooked in some prior work, as reviewed in Table \ref{tab:comparison}.

\subsection{Corpus-based Evaluation}
Corpus-based evaluation has become a widely-used type of automatic evaluation to compare different end-to-end DSs.
In the context of studying USs, it is relevant to conduct corpus-based evaluation for the following two aspects.
First, running testing of the DS trained with a specific US over a fixed testing set of dialogs could be an indirect assessment of the quality of the US.
Second, it is possible for the trained DS via RL to achieve high task success and yet not generate human language \cite{zhao-etal-2019-rethinking}, particularly when the reward is mainly defined to encourage task success.
With the fixed testing set, we could calculate BLEU which measures the NLG performance of the trained DS.

We use the standard evaluation scripts from \citet{nekvinda-dusek-2021-shades} for corpus-based evaluation. The results are shown in Table~\ref{tab:corpus} with some interesting findings.
First, the DS trained with GUS (DS-GUS) achieves higher combined score than the DS trained with ABUS (DS-ABUS). This is consistent with the results in Table~\ref{tab:cross} and again demonstrate the advantage of GUS over ABUS.
Second, note that DS-GUS is initialized from DS-SL and further trained via RL on interaction with GUS, and Table ~\ref{tab:cross} shows that DS-GUS improves over DS-SL not only in Inform and Success but also in BLEU.
This result indicates that RL training of the DS with GUS does not suffer from the tradeoff problem between policy learning and NLG in offline RL \cite{zhao-etal-2019-rethinking}\footnote{This problem for offline RL is further studied and alleviated in \citet{lubis2020lava}.}, achieving higher success and being faithful to human language.
See more discussions in Section~\ref{sec:policy}.




\begin{table}[t]
\centering
\resizebox{\linewidth}{!}{
\begin{tabular}{l|cccc}
\hline
DS &Inform &Success &BLEU &Combined\\
\hline
AuGPT \cite{kulhanek2021augpt} &76.6 &60.5 &16.8 &85.4\\
SOLOIST \cite{peng2020etal} &82.3 &72.4 &13.6 &90.9\\
UBAR \cite{yang2021ubar} &83.4 &70.3 &17.6 &94.4\\
\hdashline
DS-SL &84.10 &72.10  &19.24  &97.34\\
DS-ABUS$_{best}$ &84.20  &71.00  &19.44 &97.04\\
DS-ABUS$_{avg}$ &85.37  &69.70  &19.10  &96.64\\
DS-GUS$_{best}$ &85.70  &74.60  &19.80  &99.95\\
DS-GUS$_{avg}$ &85.17 &73.33 &19.83 &99.01\\
\hline
\end{tabular}
}
\caption{Corpus-based evaluation. Above the dashed line are GPT-2-based results from the official website of MultiWOZ.
Below are the results from DS-SL and the DSs trained with ABUS and GUS respectively.}
\label{tab:corpus}
\end{table}


\subsection{Human Evaluation}
\label{sec:human_eval}
We further perform human evaluation of the performances of USs and DSs. For each pair of US and DS, 100 dialogs were gathered, which were scored by 5 human graders. The details of evaluation metrics have been described in Sec.~\ref{sec:metrics} and the results are shown in Table~\ref{tab:human}. For convenience, we refer to the results of each row by the name of the DS in the table.
It can seen that the overall performance of DS-GUS is superior over both DS-ABUS and DS-SL.
Further, we conduct significance tests by comparing either DS-ABUS or DS-SL with DS-GUS respectively, using the matched-pairs method \cite{gillick1989some} and add a superscript $^*$ to the score in the first two rows in Table~\ref{tab:human} if the p-value is less than 0.05. All the specific p-values can be seen in Appendix~\ref{sec:sig_test}.
The results show that DS-GUS significantly improves over DS-SL for Success and US-Coh, while the differences in terms of DS-Coh, DS-Flu and US-Flu are not significant.
Moreover, all the human evaluation metrics by DS-GUS are stronger than or equal to those by DS-ABUS.
Particularly, DS-GUS significantly outperforms DS-ABUS for DS-Flu, US-Coh and US-Flu.
This indicates that GUS is able to generate more coherent and fluent utterances than ABUS. To illustrate this point, we provide some generated dialogues in Appendix~\ref{sec:case}.
\begin{table}[t]
\renewcommand\arraystretch{1.1}
\centering
\resizebox{\linewidth}{!}{
\begin{tabular}{lllllll}
\toprule
DS &US &Success &DS-Coh &DS-Flu &US-Coh &US-Flu\\
\hline
DS-ABUS &ABUS &1.71 &1.51 &1.65$^*$ &1.27$^*$ &1.30$^*$\\
DS-SL &GUS &1.73$^*$ &1.60 &1.85 &1.61$^*$ &1.88\\
DS-GUS &GUS &1.84 &1.52 &1.79 &1.75 &1.90\\
\bottomrule
\end{tabular}
}
\caption{Human evaluation of the dialogs generated by different DSs and USs. The score with $^*$ in the first two rows denotes the difference between this score and the score in the last row (DS-GUS with GUS) is significant (p-value<0.05); otherwise, the difference is not significant (p-value>=0.05).}
\label{tab:human}
\end{table}


\section{Ablation Study}
\label{sec:ablation}
\subsection{The Importance of Goal State Tracking}
In our GUS model, we use Eq.~\eqref{eq:update} to update the goal state at every turn.
In the section, we consider a variant of GUS, which sets the goal state at all turns to be the initial goal, that is, $g_t=g_0, t=1,...,T$, like in \citet{asri2016sequence, GurHUS, papangelis-etal-2019-collaborative}.
Such model is referred to as GUS w/o GST, and could be similarly trained according to Eq.~\eqref{eq:sup-us}.
Then we train a DS with this US (called ``DS-GUS w/o GST'') and test it with ABUS, GUS and GUS w/o GST respectively.
The results are shown in Table~\ref{tab:goal_ablation}. We can see that the Inform and Success rates obtained by ``DS-GUS w/o GST'' are lower than those by DS-GUS as shown in Table~\ref{tab:cross}, when testing on ABUS and GUS. This indicates the importance of using GST in GUS.
Besides, we can see that the results are pretty low when testing on GUS w/o GST. Presumably, this is because GUS w/o GST cannot accurately distinguish the uncompleted part in the complex goal, which will easily cause omission and repetition when generating user acts.
\begin{table}[t]
\centering
\resizebox{0.7\linewidth}{!}{
\begin{tabular}{l|cc}
\hline
US &Inform &Success\\
\hline
ABUS &0.863 &0.790\\
GUS &0.825 &0.777\\
GUS w/o GST &0.743 &0.502\\
\hline
\end{tabular}
}
\caption{The ablation results about goal state tracking (GST). The DS trained with GUS w/o GST is tested on ABUS, GUS and GUS w/o GST respectively.}
\vspace{-0.5em}
\label{tab:goal_ablation}
\end{table}

\subsection{Different Reward Settings}
\label{sec:reward}
The results of optimizing DS on GUS using different reward settings are reported in Table~\ref{tab:ablation1}.
It is found that all reward settings achieve better results than supervised baseline (Reward=None) and the synthetic reward setting achieves the best result, which is reasonable since the fine-grained rewards reflect more than simple success rate in terms of the nature of the tasks \cite{tseng-etal-2021-transferable}.
All RL results in this paper are based on this setting of reward, unless here for ablation study. 
\begin{table}[t]
\centering
\begin{tabular}{l|cc}
\hline
Reward &Inform &Success\\
\hline
None &0.781 &0.736\\
Success &0.842 &0.787\\
Synthetic &0.864 &0.808\\
Sigmoid synthetic &0.850 &0.780\\
\hline
\end{tabular}
\caption{Interaction-based results of testing DS-GUS on GUS under different reward settings, as introduced in Section~\ref{sec:reward}. ``None'' denotes the testing results of DS-SL with GUS, as also reported in the first row in Table~\ref{tab:cross}.}
\label{tab:ablation1}
\end{table}

\subsection{Different Policy Schemes for DS}
\label{sec:policy}
The policy in RL refers to the probabilistic mapping from states to actions.
Previous studies of learning reinforced DS, e.g., \cite{DBLP:conf/asru/LiuL17,papangelis-etal-2019-collaborative,tseng-etal-2021-transferable}, mainly employ RL to optimize the policy module, i.e., use system acts for actions.
In contrast, the policy of DS-GUS and DS-ABUS involves generating not only system act $a_t^s$, but also belief state $b_t^s$ and system response $r_t$, which can be represented as $b_t^s \oplus a_t^s \oplus r_t$, as illustrated in Eq.~\eqref{eq:rl-ds}.
To compare policy schemes for reinforced DS, we try two other policy schemes when optimizing DS-GUS. The first policy scheme only involves the generation of system act $a_t^s$ and the second one involves the generation of both system act $a_t^s$ and system response $r_t$. We denote the two policy schemes as $a_t^s$ and $a_t^s \oplus r_t$ respectively. 
Table~\ref{tab:policy} shows the interaction results when the DS-GUS trained under different policy schemes is tested with GUS.
\begin{table}
\centering
\resizebox{0.7\linewidth}{!}{
\begin{tabular}{l|cc}
\hline
Policy &Inform &Success\\
\hline
$b_t^s \oplus a_t^s \oplus r_t$ &0.864 &0.808\\
$a_t^s \oplus r_t$ &0.845 &0.770\\
$a_t^s$ &0.848 &0.796\\
\hline
\end{tabular}
}
\caption{The ablation experiments of using different policy schemes.}
\vspace{-0.9em}
\label{tab:policy}
\end{table}

It can be seen from Table~\ref{tab:policy} that using $b_t^s \oplus a_t^s \oplus r_t$ for policy achieves the highest Inform and Success rate. 
We provide two points, which may explain the advantage of our model in using $b_t^s \oplus a_t^s \oplus r_t$ for RL.
First, since the DST, DP and NLG modules in GPT-2 based DS share the model parameters, parameter adjust in one module will affect other modules. 
Only optimizing DP during RL without considering other modules may mislead other modules.
Using $b_t^s \oplus a_t^s \oplus r_t$ leads to better overall optimization and decision-making.
Second, the balance between policy learning and NLG, which was a concern in previous studies when using modular or small-capacity architectures \cite{zhao-etal-2019-rethinking}, could be relieved, thanks to the high-capacity of GPT-2.
\section{Conclusion}
In this paper, towards developing an end-to-end trainable US for multi-domains, a generative user simulator (GUS) with GPT-2 based architecture and goal state tracking is proposed and systematically evaluated.
We train GPT-2 based DSs and USs and conduct cross-model evaluation, corpus-based evaluation and human evaluation. The results show that the DS trained with GUS outperforms both the supervised trained DS and the DS trained with ABUS. The human evaluation further confirms the superiority of GUS and shows that GUS can generate much more coherent and fluent utterances than ABUS. 
Moreover, GUS is simple and easy to use, in addition to its strong performance.
Hope this work will stimulate further work on developing and using user simulators in the study of building dialog systems.
\section{Limitations}
There are some limitations of this work.
First, due to computational constraints, both the DSs and the USs are experimented based on a distilled version of GPT-2.
Studies using larger GPT-2 and other classes of larger PLMs such as T5 \cite{raffel2020t5} would enhance our results in this paper. Second, we only utilize the policy gradient method for RL in this paper. Other advanced RL methods such as proximal policy optimization (PPO) and actor-critic are also worth trying in future work.
Those being said, while we agree that experimenting with larger PLMs and more complex RL methods are meaningful, we believe the extensive experiments presented in this paper (cross-model evaluation, corpus-based evaluation, human evaluation, and ablation studies) can well support the evaluations of GUS and should not affect the main finding and contribution of this paper.
\bibliography{anthology,custom}
\bibliographystyle{acl_natbib}
\clearpage
\appendix
\section{Appendices}
\subsection{Data Processing}
\label{sec:processing}
We delexicalize system responses following \citet{zhang2020task} to reduce surface language variability. Specifically,  we replace values in the ontology with specific placeholders such as $[value\_name]$ and $[value\_price]$. The proposed DS and US are both trained on the delexicalized dataset. During human evaluation or interaction with ABUS, the system responses need to be lexicalized. We then replace those placeholders with corresponding values in the predicted entities by querying the given database with the predicted belief states.

\begin{figure}[t]
\centering
	\includegraphics[width=0.95\linewidth]{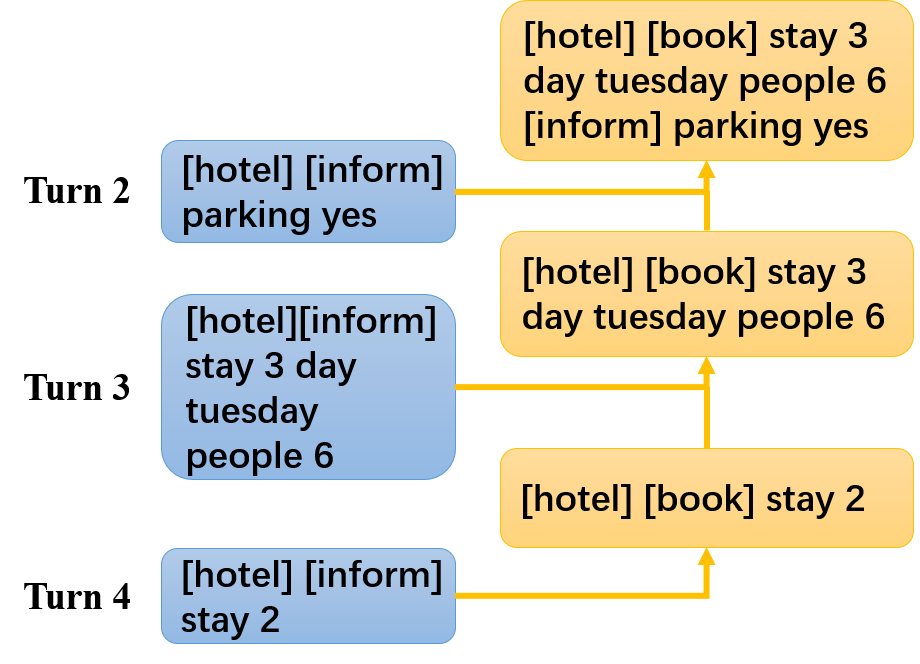}
	\caption{An example of how turn-level goal state annotations are obtained. The blue boxes are user acts and the yellow ones are goal states.}
	\vspace{-1em}
	\label{fig:goal_def}
\end{figure}

For building US, we need to accumulate the annotated user acts backwards turn by turn to obtain the goal state annotation at each turn as we described in Sec ~\ref{sec:method}. The accumulation process is depicted in Figure~\ref{fig:goal_def}.
\subsection{Implementation Details}
\label{sec:implementation}
We use Huggingface Transformers repository.
GPT-2 based DSs and USs are initialized with DistilGPT-2 \cite{sanh2019distilbert}, a distilled version of GPT-2, with 6 transformer decoder layer. During supervised learning, we use the AdamW optimizer and a linear scheduler with 20\% warm up steps and maximum learning rate $10^{-4}$. The minibatch base size is set to be 8 with gradient accumulation steps of 4.
During RL, we no longer use scheduler and fix the learning rate to $2 \times 10^{-5}$. The minibatch base size is set to be 16 with gradient accumulation steps of 12. For each interaction, the dialog will end in the following three cases: 1) both the DS and US generate \verb|bye| intent; 2) the goal state of the US is empty; 3) the content of the current turn is exactly the same as that  of the previous turn. Besides, to increase the diversity of dialogues, beam search decoding is applied during generating user acts and system acts. The beam size is set to be 10 and the final act will be sampled by probability from the 10 candidates.
All the SL and RL experiments are conducted on a single 16GB Tesla-P100 GPU.
\begin{table}[t!]
\centering
\resizebox{\linewidth}{!}{
\begin{tabular}{ll}
\toprule[1.2pt]
\multicolumn{2}{c}{\textbf{SNG0616}}\\
\hline\hline
\multirow{2}{*}{User} &Sorry, actually I need an expensive restaurant in\\
&the north. The first on your list would be great. \\
Bspan &[restaurant] pricerange expensive area north\\
Act &[restaurant] [inform] name\\
Resp &Sure how about [value\_name]?\\
\hline
Bspan$_{SL}$ &[restaurant] pricerange expensive area north food north \\
Act$_{SL}$ & [restaurant] [nooffer] food area [request] food\\
\multirow{2}{*}{Resp$_{SL}$} & I am sorry, there are no [value\_food] restaurants in the \\ &[value\_area] . Would you like to try a different type of cuisine?\\
\hline
Bspan$_{RL}$ & [restaurant] pricerange expensive area north\\
Act$_{RL}$ & [restaurant] [inform] choice price area [request] food\\
\multirow{2}{*}{Resp$_{RL}$} & There are [value\_choice] [value\_price] restaurants in\\ &the [value\_area] . What type of food would you like?\\

\bottomrule[1.2pt]
\end{tabular}
}
\caption{One dialog turn in the test set. Bspan and Act denote the sequence forms of belief state and system act. The subscript $SL$ and $RL$ represent the supervised trained model DS-SL and the RL model DS-GUS respectively.}
\label{tab:case-corpus}
\end{table}
\subsection{Case Study}
\label{sec:case}
\paragraph{Interaction Case} To illustrate the advantage of GUS over ABUS, we let DS-ABUS and DS-GUS interact with their corresponding USs under the same user goal. The generated dialogs are shown in Table~\ref{tab:case-gen}. It can be seen that the user utterances generated by ABUS are unlike natural languages and sometimes logically coherent. For instance, ABUS says \say{I want a train} when it has said \say{please book the train for me} before in the first turn. On the contrary, GUS can generates very fluent and logically coherent utterances, and is able to understand the system response and provide necessary information to complete the dialogue.
\begin{table*}[ht]
\centering
\resizebox{\linewidth}{!}{
\begin{tabular}{l}
\toprule[1.2pt]
\textbf{User Goal}\\
\textbf{Train}: Inform (arrive=12:45; day=Thursday; departure=Birmingham new street; destination: Cambridge),\\
~~~~~~~~~~~~~Book (people=8)\\
\textbf{Attraction}: Inform (area=centre), Request (postcode, phone)\\
\midrule[1.2pt]
Dialogue generated through the interaction between DS-ABUS and ABUS\\
\hline\hline
user$_1$: Please book the train for me. I am departing from Birmingham new street. I want a train leaving on Thursday.\\
~~~~~~~~~~~I would like to go to Cambridge please.\\
resp$_1$: There are 19 trains that meet your criteria. What time would you like to leave?\\
user$_2$: Please pick one for me. I don't care about the leave of the train.\\
resp$_2$: Tr4235 leaves at 05:40. Would you like me to book it for you?\\
user$_3$: I want to get there by 12:45.\\
resp$_3$: Tr4235 arrives at 08:23. Would you like me to book it for you?\\
user$_4$: I would like 8 tickets please.\\
resp$_4$: Booking was successful, the total fee is 600.8 pounds payable at the station. Reference number is: 0000002206.\\
~~~~~~~~~~~Is there anything else I can help you with?\\
user$_5$: Can you help me plan a trip to see a particular attraction? I'd like something in the centre.\\
resp$_5$: There are 44 theatre and architecture in the centre of town. What type of attraction are you interested in?\\
user$_6$: Just select one for me. I don't care about the type of the attraction.\\
resp$_6$: There are 44 museums in the centre of town. May I suggest Adc theatre ?\\
user$_7$: Sounds good. Could I get the phone number of the attraction? Can you give me the postcode of the attraction?\\
resp$_7$: The phone number is 01223300085 and the postcode is CB58AS. Is there anything else I can help you with?\\
user$_7$: That's all the help I need. Take care. Bye.\\
resp$_7$: Thank you for contacting us and have a nice day.\\
\midrule[1.2pt]
Dialogue generated through the interaction between DS-GUS and GUS\\
\hline\hline
user$_1$: I am looking for places to go in the centre of town.\\
resp$_1$: There are 44 attractions in the centre. Do you have a specific type in mind?\\
user$_2$: No, I don't have a preference.\\
resp$_2$: I would recommend Adc Theatre on Park Street. It is free to get in.\\
user$_3$: That sounds great. Can I get the phone number and postcode?\\
resp$_3$: The phone number is 01223300085 and the postcode is CB58AS. Is there anything else i can help you with?\\
user$_4$: I am also looking for a train going to Cambridge.\\
resp$_4$: Where will you be departing from and what day would you like to travel?\\
user$_5$: I will be departing from Birmingham new street on Thursday.\\
resp$_5$: There are 19 trains that meet your criteria. Do you have a preferred departure or arrival time?\\
user$_6$: I would like to arrive by 12:45. I would like to book for 8 people.\\
resp$_6$: Booking was successful, the total fee is 600.8 pounds payable at the station. Reference number is: 0000002206.\\
user$_7$: Thank you so much! that is all I need.\\
resp$_7$: Thank you for contacting us and have a nice day.\\
\bottomrule[1.2pt]
\end{tabular}
}
\caption{Dialogs generated through the interaction between DSs and USs.}
\label{tab:case-gen}
\end{table*}
\paragraph{Corpus-based Evaluation Case} We can see from Table~\ref{tab:corpus} that DS-GUS improves the Success rate over DS-SL on test set. We show an dialog example in Table~\ref{tab:case-corpus} to explain how the RL improves the DS performance. In this turn, DS-SL predicts a wrong Bspan with a redundant slot \say{food} and incorrect value \say{north}. As a result, no entity can be found when querying database and DS-SL generates \say{nooffer} intent, which finally leads to an unsuccess dialog. However, DS-GUS can predict a correct Bspan and generate an appropriate system act in this turn. This indicates that RL can improve the ability of dialog state tracking of DS, thus improving the success rate. In fact, this advantage comes from the special policy scheme employed in this paper, as discussed in Sec~\ref{sec:policy}.
\subsection{Significance Test}
\label{sec:sig_test}
In Sec.~\ref{sec:human_eval}, we conduct significance tests to show whether the differences between the first two groups and the last group in Table~\ref{tab:human} are significant. The p-values are listed in Table~\ref{tab:p-value}.
\begin{table}[t!]
\renewcommand\arraystretch{1.1}
\centering
\resizebox{\linewidth}{!}{
\begin{tabular}{lcccccc}
\toprule
DS &Success &DS-Coh &DS-Flu &US-Coh &US-Flu\\
\hline
DS-ABUS vs DS-GUS  &0.065 &0.535 &0.036 &0.000 &0.000\\
DS-SL vs DS-GUS &0.045 &0.220 &0.273 &0.020 &0.639\\
\bottomrule
\end{tabular}
}
\caption{Significance tests for human evaluation. We refer to the results of each row in Table~\ref{tab:human} by the name of the DS.}
\vspace{-0.5em}
\label{tab:p-value}
\end{table}
\end{document}